\documentclass[runningheads]{llncs}
\usepackage[T1]{fontenc}
\usepackage{xcolor}
\usepackage{graphicx}

\usepackage[T1]{fontenc}
\usepackage{graphicx}
\usepackage{makecell}

\usepackage{cite}
\usepackage{amsmath,amssymb,amsfonts}
\usepackage{algorithmic}

\usepackage{textcomp}
\usepackage{xcolor}

\usepackage{algorithm}
\usepackage{color}

\usepackage{caption}

\usepackage{subfigure}

\usepackage{marvosym}

\begin{document}
\title{Defending Collaborative Filtering Recommenders via Adversarial Robustness–Based Edge Reweighting}
%
%
\author{Yongyu Wang}
%
%
\institute{}  

\maketitle              
\renewcommand{\thefootnote}{}

\renewcommand{\thefootnote}{}
\footnotetext{Correspondence to: yongyuw@mtu.edu}

\begin{abstract}
User-based collaborative filtering (CF) relies on a user–user similarity graph, making it vulnerable to profile injection (shilling) attacks that manipulate neighborhood relations to promote (push) or demote (nuke) target items. In this work, we propose an adversarial robustness–based edge reweighting defense for CF. We first assign each user–user edge a non-robustness score via spectral adversarial robustness evaluation, which quantifies the edge’s sensitivity to adversarial perturbations. We then attenuate the influence of non-robust edges by reweighting similarities during prediction. Extensive experiments demonstrate that the proposed method  effectively defends against various types of attacks.

\end{abstract}
\section{Introduction}

Collaborative filtering (CF) is a core method in recommender systems\cite{su2009survey,koren2009matrix,he2020lightgcn}. It supports personalized recommendation and has been applied to multiple domains\cite{smith2017two,gomez2015netflix}. However, because recommender systems typically accept large-scale user feedback from open environments, they are vulnerable to attacks and deception. For instance, adversaries may tamper with genuine rating data or inject fabricated ratings to conduct (i) \emph{push attacks}, which aim to promote a target item by increasing its predicted score, ranking, or recommendation frequency, and (ii) \emph{nuke attacks}, which seek to demote a target item by decreasing these quantities \cite{gunes2014shilling}.

Existing defenses against attacks in collaborative filtering (CF) generally fall into two lines of research~\cite{zhang2025robust}: (i) identifying malicious (fake) users and reducing their influence~\cite{zhang2025robust,zhang2014hht,yang2016re,zhang2020gcn,zhang2024lorec,liu2020recommending}, and (ii) improving model robustness through adversarial training, commonly referred to as Adversarial Collaborative Filtering (ACF)~\cite{he2018adversarial,chen2019adversarial,li2020adversarial,wu2021fight,ye2023towards}. 
The former line either filters out suspicious users prior to model learning~\cite{zhang2025robust,zhang2014hht,yang2016re,liu2020recommending} or suppresses their contribution during training~\cite{zhang2020gcn,zhang2024lorec}. 
In practice, such approaches typically depend on hand-crafted assumptions about attack behaviors~\cite{zhang2025robust,zhang2020gcn} or require attack-related supervision (i.e., labeled malicious data)~\cite{zhang2014hht,yang2016re,zhang2020gcn,zhang2024lorec}. 
As a result, when real-world attacks deviate from these assumed patterns, the detector may misclassify users, which not only weakens the defense but can also degrade the experience of benign users~\cite{zhang2024lorec}.

In contrast, adversarial collaborative filtering (ACF) tackles poisoning threats without assuming a specific attack recipe \cite{he2018adversarial,chen2019adversarial,li2020adversarial,wu2021fight,ye2023towards}. A representative line of ACF methods is built upon the Adversarial Personalized Ranking (APR) framework \cite{he2018adversarial}, which improves robustness by explicitly regularizing the learned user and item embeddings. Since poisoning attacks often manifest as shifts in these embeddings \cite{huang2021data,tang2020revisiting}, APR-style defenses inject bounded adversarial perturbations into the embedding space during training and optimize a min–max objective, training the recommender to perform well under worst-case embedding perturbations \cite{he2018adversarial,li2020adversarial,ye2023towards}. However, it has been observed that adversarial training can degrade model performance on clean samples~\cite{tramer2019adversarial,weng2020trade}. Moreover, several studies have theoretically demonstrated the existence of a trade-off between robustness against evasion attacks and the performance of adversarial training~\cite{zhang2019theoretically}.

To address the above limitations, we propose an adversarial robustness–based edge reweighting scheme to defend user-based collaborative filtering (CF) recommenders. We first assign each user–user edge a non-robustness score via a spectral adversarial robustness evaluation procedure, which quantifies how sensitive the neighborhood relation is to worst-case perturbations. We then attenuate the influence of non-robust connections by reweighting similarity edges during rating prediction, so that unreliable neighbors contribute less to the final estimate. In this way, our method preserves the utility of CF on benign data while substantially reducing the impact of profile injection (shilling) attacks that attempt to promote or demote target items.

\section{BACKGROUND}

\subsection{Collaborative Filtering Recommendation Algorithms}

Given a user set $U=\{u_1,\ldots,u_n\}$ and an item set $I=\{i_1,\ldots,i_m\}$, we form an $n\times m$ user--item interaction matrix by collecting historical feedback such as ratings, reviews, or purchase records. Each row corresponds to a user and each column corresponds to an item; an entry stores the observed rating (or another interaction signal) between the corresponding user and item. We refer to the target user as the \emph{active user} $u_a$, for whom the recommender system needs to predict the rating of a target item. A standard user-based CF algorithm consists of two main steps:

\textbullet\ \emph{Neighbor selection.} The algorithm computes the similarity (or distance) between $u_a$ and other users based on their interaction vectors, and selects the $k$ most similar users (or equivalently, the $k$ nearest users under a distance metric) as the neighborhood of $u_a$. Common choices include cosine similarity and Euclidean distance.

\textbullet\ \emph{Rating prediction.} Let the $k$ nearest neighbors of $u_a$ be denoted by $nn_1,\ldots,nn_k$ with corresponding similarity scores $s_1,\ldots,s_k$. For a target item $i_a$ whose rating for $u_a$ is to be predicted, let $r_1,\ldots,r_k$ denote the ratings of $i_a$ provided by these neighbors (typically restricted to neighbors who have rated $i_a$). The predicted rating $P_{u_a,i_a}$ is computed as a similarity-weighted average:
\begin{equation}\label{eqn:weightedsum}
P_{u_a,i_a} = \frac{\sum_{i=1}^{k} s_i \cdot r_i}{\sum_{i=1}^{k} s_i}.
\end{equation}
When the denominator is zero or no neighbor has rated $i_a$, a common practice is to fall back to a user/item/global mean.

\subsection{Spectral Graph Theory}

Spectral graph theory analyzes graph structure via the eigenvalues and eigenvectors of matrices associated with the graph, most commonly the graph Laplacian. Consider a weighted graph \( G = (V, E, w) \), where \( V \) denotes the set of vertices, \( E \) denotes the set of edges, and \( w \) assigns a positive weight to each edge. The Laplacian matrix of \( G \), which is symmetric and diagonally dominant (SDD), is defined as
\begin{equation}\label{formula_laplacian}
L(p,q) = 
\begin{cases}
-w(p,q) & \text{if } (p,q) \in E, \\
\sum\limits_{(p,t) \in E} w(p,t) & \text{if } p = q, \\
0 & \text{otherwise}.
\end{cases}
\end{equation}
To improve numerical stability and reduce sensitivity to scaling, it is often preferable to work with the normalized Laplacian
\[
L_{\text{norm}} = D^{-1/2} L D^{-1/2},
\]
where \( D \) is the diagonal degree matrix.

In many graph learning and spectral analysis tasks, only a limited number of eigenvalues and eigenvectors are needed~\cite{chung1997spectral}. For example, when one is interested in revealing community or clustering structure, the smallest Laplacian eigenvalues and their corresponding eigenvectors are typically the most informative~\cite{von2007tutorial}. As a result, computing a full eigendecomposition is unnecessary and becomes prohibitively expensive for large graphs. The Courant–Fischer minimax theorem~\cite{golub2013matrix} offers a variational view that enables iterative computation of eigenvalues without explicitly computing the entire spectrum. Specifically, the \(k\)-th eigenvalue of a Laplacian matrix \( L \in \mathbb{R}^{|V| \times |V|} \) can be written as
\[
\lambda_{k}(L) = \min_{\dim(U) = k} \max_{x \in U, \, x \neq 0} \frac{x^T L x}{x^T x}.
\]

In practice, one often needs to compare two matrices or transfer information between them, rather than studying a single matrix in isolation. In this scenario, the Courant–Fischer theorem admits a generalized extension. Let \( L_X \in \mathbb{R}^{|V| \times |V|} \) and \( L_Y \in \mathbb{R}^{|V| \times |V|} \) be two Laplacian matrices satisfying \( \text{null}(L_Y) \subseteq \text{null}(L_X) \). For \( 1 \leq k \leq \text{rank}(L_Y) \), the \(k\)-th eigenvalue of the matrix \( L_Y^+ L_X \) has the following variational characterization:
\[
\lambda_k(L_Y^+ L_X) = 
\min_{\substack{\dim(U) = k,\\ U \perp \text{null}(L_Y)}} \ 
\max_{x \in U,\, x \neq 0} \frac{x^T L_X x}{x^T L_Y x},
\]
where \(L_Y^+\) denotes the Moore–Penrose pseudoinverse of \(L_Y\).

\section{Method}
\label{sec:method}

The core idea of our defense is to distinguish between user--user edges that remain reliable under perturbations (robust edges) and those that are highly sensitive to profile manipulation (non-robust edges). Since user-based collaborative filtering (CF) makes predictions by propagating information through a user--user similarity graph, an attacker can distort recommendations by injecting or altering rating profiles, thereby reshaping neighborhoods and amplifying the influence of malicious connections. To mitigate this vulnerability, we assign each user--user edge a quantitative non-robustness score using a spectral adversarial robustness evaluation method (Spade), and then attenuate the contribution of non-robust edges during CF prediction. The proposed framework consists of the following four phases.

\subsection{Phase 1: User Similarity Graph Construction}
User-based CF relies on a user--user similarity graph constructed from historical interactions (training data). Each node corresponds to a user, and each weighted edge encodes the similarity between two users computed from their rating vectors (e.g., cosine similarity). Following common practice in neighborhood-based CF, we build a weighted $k$-nearest neighbor ($k$-NN) graph by connecting each user to its $k$ most similar users, and symmetrize the graph to obtain an undirected weighted adjacency matrix, denoted by $G_X=(V,E_X,w)$.

This graph serves as the prediction backbone for CF, but it also becomes the main attack surface: manipulating user profiles can create spurious high-similarity edges, causing the model to over-trust compromised neighborhoods. In constructing $G_X$, we adopt three stabilizers consistent with neighborhood CF practice: (i) mean-centering is applied \emph{only on observed entries} when forming the feature matrix used for similarity computation, (ii) negative similarities are clipped to zero, and (iii) a small positive safeguard is applied when a user’s top-$k$ similarities are all non-positive, to avoid isolated nodes and unstable neighborhoods.

\subsection{Phase 2: Reference Manifold Construction in an Embedding Space}
Spade evaluates edge stability by comparing the input graph to a reference manifold graph constructed in an embedding space. Intuitively, the reference graph captures a ``clean'' geometric structure of user representations, and edges that disagree with this structure tend to be fragile under perturbations.
Concretely, we compute a user embedding matrix and then build an auxiliary $k$-NN graph in that embedding space using $k_{\text{ref}}$ neighbors per node, denoted by $G_Y=(V,E_Y)$. In our implementation, the embedding can be obtained via (i) spectral embedding of the Laplacian of $G_X$, or (ii) a rating-based embedding such as truncated SVD on the user--item interaction matrix. The latter provides a less ``same-source'' reference because it does not reuse the original user graph topology, and empirically leads to more stable robustness estimates.

\subsection{Phase 3: Spectral Adversarial Robustness Evaluation (Spade)}
Let \(L_X\) and \(L_Y\) denote the Laplacian matrices of the original user similarity graph and the auxiliary $k$-NN graph constructed in the embedding space, respectively. Following the spectral robustness framework in~\cite{cheng2021spade}, we use dominant generalized eigenvalues and eigenvectors associated with \(L_Y^+ L_X\) to quantify how sensitive each user--user connection is to perturbations.
Specifically, we form a weighted eigensubspace matrix \(V_s \in \mathbb{R}^{|V| \times s}\) as
\[
V_s =
\begin{bmatrix}
v_1 \sqrt{\zeta_1}, & v_2 \sqrt{\zeta_2}, & \ldots, & v_s \sqrt{\zeta_s}
\end{bmatrix},
\]
where \(\zeta_1 \geq \zeta_2 \geq \cdots \geq \zeta_s\) are the largest \(s\) eigenvalues of \(L_Y^+ L_X\), and \(v_1, v_2, \ldots, v_s\) are the corresponding eigenvectors.

Each user \(p\) is represented by the \(p\)-th row of \(V_s\). For an edge \((p,q)\) in the original user graph, its non-robustness is measured by the spectral embedding distance between its endpoints:
\[
\| V_s^\top e_{p,q} \|_2^2,
\]
where \(e_{p,q}\) is the incidence vector of edge \((p,q)\). The \textit{Spade score} is defined as
\[
\text{Spade}(p, q) = \| V_s^\top e_{p,q} \|_2^2.
\]
Edges with larger \(\text{Spade}(p,q)\) values are considered less robust, meaning that small perturbations in user profiles can more easily change their effective influence in neighborhood-based prediction.

\subsection{Phase 4: CF Defense via Edge Reweighting (and Optional Pruning)}
Given Spade scores on the user graph, we defend CF by reducing the influence of non-robust edges during prediction. Instead of treating all similarity edges as equally reliable, we compute a robustness-aware similarity weight
\[
w'(p,q) = \frac{w(p,q)}{1 + \gamma \cdot \widetilde{\text{Spade}}(p,q)},
\]
where \(w(p,q)\) is the original similarity weight, \(\widetilde{\text{Spade}}(p,q)\) is a normalized Spade score (e.g., scaled to $[0,1]$), and $\gamma \ge 0$ controls the strength of attenuation. CF then uses $w'(p,q)$ for neighbor selection and/or similarity-weighted aggregation.

Optionally, a pruning variant can be used by removing a fraction of edges with the largest Spade scores. However, for recommendation tasks, aggressive pruning may increase the fallback rate (i.e., no neighbor rated the target item), so we primarily focus on reweighting, which preserves connectivity while suppressing suspicious edges.

\begin{algorithm}[!t]
\caption{Robustness-Weighted User-based CF}
\label{alg:spade_cf}
\textbf{Input:} Training interactions (ratings) $R$, $k_{\text{graph}}$, $k_{\text{spec}}$, $s$, $\gamma$.\\
\textbf{Output:} Robust CF predictions (or top-$N$ recommendations).\\
\begin{algorithmic}[1]
    \STATE Construct a user feature matrix from $R$ for similarity computation (e.g., mean-centering only on observed entries).
    \STATE Build a weighted user--user $k_{\text{graph}}$-NN graph $G_X=(V,E_X,w)$ using cosine similarity.
    \STATE Compute a user embedding (e.g., truncated SVD on $R$ or spectral embedding of $G_X$).
    \STATE Build an auxiliary unweighted $k_{\text{spec}}$-NN graph $G_Y$ in the embedding space and form its Laplacian $L_Y$.
    \STATE Form the Laplacian $L_X$ of $G_X$.
    \STATE Compute dominant generalized eigenpairs of \(L_Y^+ L_X\) (or an equivalent regularized generalized eigenproblem) and construct \(V_s\).
    \STATE For each \((p,q)\in E_X\), compute \(\text{Spade}(p,q)=\|V_s^\top e_{p,q}\|_2^2\) and normalize to \(\widetilde{\text{Spade}}(p,q)\).
    \STATE Reweight user similarities by \(w'(p,q)=\frac{w(p,q)}{1+\gamma\cdot \widetilde{\text{Spade}}(p,q)}\).
    \STATE Perform user-based CF prediction using the robustness-aware weights $w'$ (for neighbor selection and aggregation).
\end{algorithmic}
\end{algorithm}


\section{Experiments}
\label{sec:experiments}

\subsection{Experimental Setup}

\paragraph{Dataset and split.}
We conduct experiments on MovieLens-100K, which contains 100,000 explicit ratings from 943 users on 1,682 items.
We use a per-user holdout split: for each user, a fixed fraction of interactions is held out for testing and the remaining interactions are used for training.
All results are reported as mean $\pm$ standard deviation over multiple random seeds.

\paragraph{Backbone recommender.}
Our backbone is a classical user-based neighborhood CF model.
It builds a weighted user--user similarity graph from training interactions and predicts ratings by similarity-weighted neighborhood aggregation.
Our defense is applied on top of the same CF backbone under identical train/test splits.

\paragraph{Attack setting.}
We evaluate profile injection (shilling) attacks, where an adversary injects malicious user profiles into the training data to manipulate recommendations.
We focus on \emph{push}-style attacks that aim to promote a target item by creating spurious neighborhood connections (e.g., by co-rating popular items to blend in).
For each run, we sample a target item and measure how much the attack increases its predicted preference and recommendation exposure among users who have not interacted with it.

\paragraph{Defense protocol.}
To evaluate whether robustness-aware edge reweighting can mitigate profile manipulation, we compute edge non-robustness scores once from the pre-attack user similarity graph and then apply the resulting reweighting when making predictions under attack.

\subsection{Evaluation Metrics}

\paragraph{Rating prediction accuracy.}
We report Mean Absolute Error (MAE) and Root Mean Squared Error (RMSE) on the test set.

\paragraph{Attack impact on a target item.}
We form a candidate user set consisting of users who have not rated the target item, and report:
(i) \textbf{Target-score uplift}, the increase of the mean predicted score of the target item caused by the attack;
(ii) \textbf{HR@N uplift}, the increase of the hit rate of the target item appearing in the top-$N$ list (with a fixed $N$);
(iii) \textbf{MeanRank gain}, the average improvement in the target item’s rank (clean rank minus attacked rank; larger means the item moves higher).
For each metric, we also report \textbf{Reduction}, defined as the decrease of the attack effect achieved by the defense relative to the original CF (larger is better).

\subsection{Results and Discussion}

Table~\ref{tab:acc_atk} reports rating prediction accuracy under attacked training data.
The proposed robustness-aware edge reweighting preserves (and slightly improves) MAE/RMSE compared to the original CF backbone, indicating that robustness gains do not come at the expense of predictive accuracy.

Table~\ref{tab:attack_effect} summarizes how the attack manipulates the target item.
The original CF exhibits substantial increases in the target item’s predicted score and exposure (HR@N), as well as a large rank improvement.
In contrast, robustness-aware reweighting consistently suppresses these effects, yielding markedly smaller uplifts and rank gains.
Overall, the results support that down-weighting non-robust user--user edges can reduce the propagation of malicious neighborhood influence under profile injection attacks.

\begin{table*}[!t]
\centering
\caption{Rating prediction performance (mean $\pm$ std) under attacked training data. Lower is better.}
\label{tab:acc_atk}
\setlength{\tabcolsep}{10pt}
\small
\begin{tabular}{lrr}
\hline\hline
Method &
\makecell{Attacked\\MAE} &
\makecell{Attacked\\RMSE} \\
\hline
Original CF & 0.7356 $\pm$ 0.0023 & 0.9431 $\pm$ 0.0041 \\
Ours        & 0.7349 $\pm$ 0.0024 & 0.9419 $\pm$ 0.0043 \\
\hline\hline
\end{tabular}
\end{table*}

\begin{table*}[!t]
\centering
\caption{Attack impact on the target item (mean $\pm$ std). ``Reduction'' is the decrease in attack effect achieved by the defense (larger is better).}
\label{tab:attack_effect}
\setlength{\tabcolsep}{10pt}
\small
\begin{tabular}{lrrr}
\hline\hline
Metric &
\makecell{Original CF\\(attack effect)} &
\makecell{Ours\\(attack effect)} &
\makecell{Reduction\\(Ours vs. Original)} \\
\hline
Target-score uplift & 0.5669 $\pm$ 0.0773 & 0.3246 $\pm$ 0.0496 & 0.2422 $\pm$ 0.0356 \\
HR@N uplift         & 0.2652 $\pm$ 0.0485 & 0.1052 $\pm$ 0.0398 & 0.1600 $\pm$ 0.0105 \\
MeanRank gain       & 272.42 $\pm$ 21.99  & 163.34 $\pm$ 15.45  & 109.08 $\pm$ 14.63 \\
\hline\hline
\end{tabular}
\end{table*}

\subsection{Efficiency}
By pruning non-robust edges in the graph, our method can perform prediction by using only a small subset of neighbors: for each query $(u,i)$,  which lowers memory usage and accelerates inference by shortening per-user neighbor scans. This leads to a favorable accuracy--efficiency trade-off: substantial sparsification can be achieved without degradation in prediction accuracy.

\section{Conclusion}
We studied the vulnerability of user-based collaborative filtering to profile injection attacks and proposed an adversarial robustness--based edge reweighting strategy that down-weights non-robust user--user relations identified by a spectral robustness evaluation. Experiments show that the proposed method substantially reduces the attack-induced promotion of target items while preserving standard rating prediction accuracy. These results suggest that robustness-aware treatment of neighborhood edges is a simple yet effective direction for defensing collaborative filtering recommenders.


%
%
%
%

\end{document}